\documentclass[11pt,a4paper]{article}
\usepackage[nohyperref]{naaclhlt2019}
\usepackage{times}
\usepackage{latexsym}

\usepackage{times}
\usepackage{epsfig}
\usepackage{graphicx}
\usepackage{amsmath}
\usepackage{amssymb}
\usepackage{footmisc}
\usepackage{multirow}
\usepackage{caption}
\usepackage{subcaption}
\usepackage{booktabs}

\def\x{{\mathbf x}}
\def\X{{\mathbf X}}
\aclfinalcopy 

\newcommand\BibTeX{B{\sc ib}\TeX}

\title{An Effective Label Noise Model for DNN Text Classification}

\author{Ishan Jindal\textsuperscript{1}, Daniel Pressel\textsuperscript{2}, Brian Lester\textsuperscript{2}, and Matthew Nokleby\textsuperscript{1} \\
\textsuperscript{1}Wayne State University, Detroit MI 48202\\
\textsuperscript{2}Interactions Digital Roots, Ann Arbor MI 48104\\ 
  {\tt \{ishan.jindal, matthew.nokleby\}@wayne.edu},\\ {\tt \{dpressel, blester\}@interactions.com}}
\date{}
\usepackage{etoolbox}
\makeatletter
\patchcmd\@combinedblfloats{\box\@outputbox}{\unvbox\@outputbox}{}{\errmessage{\noexpand patch failed}}
\makeatother
\begin{document}
\maketitle

\begin{abstract}
Because large, human-annotated datasets suffer from labeling errors, it is crucial to be able to train deep neural networks in the presence of label noise. While training image classification models with label noise have received much attention, training text classification models have not. In this paper, we propose an approach to training deep networks that is robust to label noise. This approach introduces a non-linear processing layer (\emph{noise model}) that models the statistics of the label noise into a convolutional neural network (CNN) architecture. The noise model and the CNN weights are learned jointly from noisy training data, which prevents the model from overfitting to erroneous labels. Through extensive experiments on several text classification datasets, we show that this approach enables the CNN to learn better sentence representations and is robust even to extreme label noise. We find that proper initialization and regularization of this noise model is critical. Further, by contrast to results focusing on large batch sizes for mitigating label noise for image classification, we find that altering the batch size does not have much effect on classification performance.
\end{abstract}


\section{Introduction}
\label{sec:introduction}
Deep Neural Networks (DNNs) have led to significant advances in the fields of computer vision \cite{he2016deep}, speech processing \cite{graves2013speech} and natural language processing \cite{kim2014convolutional,young2018recent,devlin2018bert}.
To be effective, supervised DNNs rely on large amounts of carefully labeled training data. However, it is not always realistic to assume that example labels are clean. Humans make mistakes and, depending on the complexity of the task, there may be disagreement even among expert labelers. To support noisy labels in data, we need new training methods that can be used to train DNNs directly from the corrupted labels to significantly reduce human labeling efforts.
Zhu and Wu \shortcite{zhu2004class} perform an extensive study on the effect of label noise on classification performance of a classifier and find that noise in input features is less important than noise in training labels. 

In this work, we add a \emph{noise model} layer on top of our target model to account for label noise in the training set, following \cite{jindal2016learning,sukhbaatar2014training}. We provide extensive experiments on several text classification datasets with artificially injected label noise. We study the effect of two different types of label noise; \emph{Uniform label flipping (Uni)}, where a clean label is swapped with another label sampled uniformly at random; and \emph{ Random label flipping (Rand)} where a clean label is swapped with another label from the given number of labels sampled randomly over a unit simplex.

We also study the effect of different initialization, regularization, and batch sizes when training with noisy labels. We observe that proper initialization and regularization helps the noise model learn to be robust to even extreme amounts of noise. Finally, we use low-dimensional projections of the features of the training examples to understand the effectiveness of the noise model.

The rest of the paper is organized as follows. Section \ref{sec:literature} discusses the various approaches in literature to handle label noise. In Section \ref{sec:problem}, we describe the problem statement along with the proposed approach. We describe the experimental setup and datasets in Section \ref{sec:experiment}. We empirically evaluate the performance of the proposed approach along with the discussion in Section \ref{sec:results} and finally conclude our work in Section \ref{sec:conclusion}.


\section{Related Work}
\label{sec:literature}

Learning from label noise is a widely studied problem in the classical machine learning setting. Earlier works \cite{brodley1999identifying,rebbapragada2007class,manwani2013noise} consider learning from noisy labels for a wide range of classifiers including SVMs \cite{natarajan2013learning} and fisher discriminants \cite{lawrence2001estimating}. Traditional approaches handle label noise by detecting and eliminating the corrupted labels. More details about these approaches can be found in \cite{frenay2014classification}.

Recently, DNNs have made huge gains in performance over traditional methods on large datasets with very clean labels. However large real-world datasets often contain label errors. A number of works have attempted to address this problem of learning from corrupted labels for DNNs. These approaches can be divided into two categories; attempts to mitigate the effect of label noise using auxiliary clean data, and attempts to learn directly from the noisy labels. 

\textbf{Presence of auxiliary clean data:} This line of research exploits a small, clean dataset to correct the corrupted labels. For instance, \cite{li2017learning} learn a teacher network with clean data to re-weight a noisy label with a soft label in the loss function. Similarly, \cite{veit2017learning} use the clean data as a label correction network. One can use this auxiliary source of information to do inference over latent clean labels \cite{vahdat2017toward}. Further, \cite{yao2018deep} models the auxiliary trustworthiness of noisy image labels to alleviate the effect of label noise. Though these methods show very promising results, the absence of clean data in some situations might hinder the applicability of these methods. 

\textbf{Learning directly from noisy labels:} This research directly learns from the noisy labels by designing a robust loss function, or by modeling the latent labels. For instance, \cite{reed2014training}, apply bootstrapping to the loss function to have consistent label prediction for similar images. Similarly, \cite{joulin2016learning} alleviate the label noise effect by adequately weighting the loss function using the sample number. \cite{jiang2017mentornet} propose a sequential meta-learning model that takes in a sequence of loss values and outputs the weights for the labels. \cite{ghosh2017robust} further explores the conditions on loss functions such that the loss function is noise tolerant. 

A number of approaches learn the transition from latent labels to the noisy labels. For example, \cite{mnih2012learning} propose a noise adaptation framework for symmetric label noise. Based on this work, several other works \cite{sukhbaatar2014training,jindal2016learning,patrini2017making,han2018masking} account for the label noise by learning a noisy layer on top of a DNN where the learned transition matrix represents the label flip probabilities. Similarly, \cite{xiao2015learning} propose a probabilistic image conditioned noise model. \cite{azadi2015auxiliary} proposed an image regularization technique to detect and discard the noisy labeled images. Other approaches include building two parallel classifiers \cite{misra2016seeing} where one classifier deals with image recognition and the other classifier models human’s reporting bias.

All of these approaches have targeted image classification. In this work, we propose a framework for learning from noisy labels for text classification using a DNN architecture. Similar to \cite{sukhbaatar2014training,jindal2016learning,patrini2017making}, we append a non-linear processing layer on top of this architecture to model the label noise. This layer helps the base architecture to learn better representations, even in the presence of label noise. We empirically show that, for better classification performance, the knowledge of noise transition matrix is not needed. Instead, the process forces the DNN to learn better sentence representations.


\section{Problem Statement}
\label{sec:problem}

In a supervised text classification setting where $\x_i \in \mathbf{R}^d$ is a $d$-dimensional word embedding of the $i$th word in a sentence of length $T$ (padded wherever necessary), we represent the sample as an temporal embedding matrix $\X \in \mathbf{R}^{d\times T}$ which belongs to one of the $K$ classes. Let the noise-free training set be denoted by
	$\mathcal{D} = \{ (\X_1,y_1), (\X_2, y_2), \cdots, (\X_n,y_n)\},$
where $y_i \in \{1,\dots,K\}$ represents the category of $i$th sample, $n$ is the total number of training samples, and there is an unknown joint distribution $p(\X,y)$ on the sample/label pairs. This temporal representation of a sample $\X$ is fed as input to a classifier on the training set $\mathcal{D}$ with sample categories $y$. However, as mentioned in Section \ref{sec:literature}, we cannot access the true noise-free samples labels and instead, observe noisy labels corrupted by an unknown noise distribution. Let this {\em noisy} training set be denoted by 
\begin{equation*}
	\mathcal{D}^\prime = \{ (\X_1,y_1^\prime), (\X_2, y_2^\prime), \cdots, (\X_n,y_n^\prime)\},
\end{equation*}
where $y_i^\prime$ represents the corrupted label for the sentence $\X_i$. In this work, we suppose the label noise is {\em class-conditional}, where the noisy label $y_i^\prime$ depends only on the true label $y_i$, but not on the input $\X_i$ or any other labels $y_j$ or $y_j^\prime$. Under this model, the label noise is characterized by the conditional distribution $(y^\prime = i| y = j) = \Phi_{ij},$ which we describe via the $K\times K$ column-stochastic matrix $\Phi_{ij}$, parameterized by a matrix $\mathbf{Q} = \{\Phi_{ij}\}$.

In our experiments, we artificially inject label noise into the training and validation sets. We fix the noise distribution $\Phi_{ij}$ and, for a training sample, we generate a noisy label by drawing i.i.d from this noise distribution $\Phi_{ij}$. However, we do not alter the test labels. 

Though the proposed approach works for any noise distribution, for this study, we focus on two different types of label flip distributions. We use a noise model parameterized by the overall probability of a label error, denoted by $0 \leq p \leq 1$. For a noise level $p$, we set the noise distribution matrix
\begin{equation}
\Phi = (1-p)\mathbf{I} + \frac{p}{K}\mathbf{II}^K,
\end{equation}
and we call it \emph{Uniform label flip noise model}. Here, $\mathbf{I}$ represents the identity matrix and $\mathbf{II}$ denotes the all-ones matrix. Similarly, we describe the \emph{random label flip noise model} as 
\begin{equation}
	\Phi = (1-p)\mathbf{I} + p \Delta,
	\label{eq::ProblemDef::NoiseModel}
\end{equation}
where $\mathbf{I}$ is the identity matrix, and $\Delta$ is a matrix with zeros along the diagonal and remaining entries of each column are drawn uniformly and independently from the $K-1$-dimensional unit simplex. The label error probability for each class is $p$, while the probability distribution {\em within} the erroneous classes is drawn uniformly at random.

Our objective is to train a classifier on the noisy labeled sample categories on the training set $\mathcal{D}^\prime$ such that it jointly makes accurate predictions of the true label $y$ and learns the noise transition matrix simultaneously, given $\X$. For the noisy dataset $\mathcal{D}^\prime$, it is straightforward to train a classifier that predicts the {\em noisy} labels using conditional distribution for the noisy labeled input sentence $\X$:
\begin{multline}
p(y^\prime = \hat{y}^\prime|\X) = \sum_i \Big(p(y^\prime = \hat{y}^\prime|y = \hat{y}_i)\\ p(y = \hat{y}_i|\X)\Big).
\label{eq::model_eq_exp}
\end{multline}
One can learn the classifier associated with $p(y^\prime = \hat{y}^\prime|x)$ via standard training on the noisy set $\mathcal{D}^\prime$. To predict the clean labels by learning the conditional distribution $p(y = \hat{y}_i|x)$ requires more effort, as we cannot extract the ``clean'' classifier from the noisy classifier when the label noise distribution is unknown.

\subsection{Proposed Framework}
We refer to the DNN model without the final layer as the \emph{base model} or network \emph{without noise model (WoNM)}. This model, along with the non-linear layer, is trained via back-propagation on the noisy training dataset. The non-linear processing layer in the \emph{noise model} transforms the base model outputs to match the noisy labels during the forward pass better and presents the denoised labels to the base model during the backward pass. The noise layer is parameterized by a square matrix $\Psi \in \mathbf{R}^{K\times K}$). At test time, we remove this learned noise model and use the output of the base model as final predictions.

We refer to the base model parameters as $\Theta$. The base model outputs a probability distribution over the number of $K$ categories denoted as $p\left(y=\hat{y}_i|\X;\Theta \right) \forall i \in \{1, 2, \cdots, K\}$. During the forward pass the noise model transforms this output to obtain the noisy labels as
\begin{equation}
p\left(y^{\prime}|\X; \Theta, \Psi \right) = \sigma\left(\Psi \times p\left(y|\X;\Theta \right)\right),
\label{eq::model_eq}
\end{equation}
where $\sigma(\cdot)$ represents the usual softmax operator. Note that both the equations \eqref{eq::model_eq_exp} and \eqref{eq::model_eq} compute the probability distribution over noisy labels -- our noise model does not learn a noise transition matrix. However, we assert that the knowledge of exact noise statistics is neither necessary nor sufficient for the better prediction results.

We learn the base model parameters $\Theta$ and the noise model parameters $\Psi$ by maximizing the log likelihood \eqref{eq::model_eq} over all of the training samples, minimizing the \emph{cross-entropy} loss:
\begin{align}
\mathcal{L}(\Theta, \Psi;\mathcal{D}^{\prime}) &= - \frac{1}{n}\sum_{i=1}^n \log \left[p\left(y^{\prime}|\X_i; \Theta, \Psi \right)\right] \nonumber \\
&=- \frac{1}{n}\sum_{i=1}^n \log \left[\sigma\left(\Psi \times p\left(y|\X_i;\Theta \right)\right)\right]_{y_i}
\end{align} 
Similar to \cite{sukhbaatar2014training}, we initialize the noise model weights to the identity matrix. Since DNNs have high capacity, we may encounter the situation when the model absorbs all the label noise and, thus, the noise model does not learn anything at all. In order to avoid this situation, and to prevent overfitting, we apply $l_2$ regularization to the noise model. However, we want the noise model to overfit the label noise. In the experiment section, we observe that with proper regularization and weight initialization the noise model absorbs most of the label noise. Finally, we train the entire network according to the following loss function:
\begin{multline}
\mathcal{L} = - \frac{1}{n}\sum_{i=1}^n \log \left[\sigma\left(\Psi \times p\left(y|\X_i;\Theta \right)\right)\right]_{y_i} \\+ \frac{1}{2}\lambda||\Psi||_2^2.
\end{multline}
Here, $\lambda$ is a tuning parameter and we validate the value of $\lambda$ by repeating the experiment multiple times with multiple $\lambda$ values over different datasets and choose the one with better classification performance. A value of $\lambda=0.01$ works best.



\section{Datasets and Experimental Setup}
\label{sec:experiment}

In this section, we empirically evaluate the performance of the proposed approach for text classification and compare our results with the other methods. 

\subsection{General Setting}
In all the experiments, we use a publicly-available deep learning library \emph{Baseline} -- a fast model development tool for NLP tasks \cite{W18-2506}. For all the different datasets, we choose a commonly-used, high-performance model from \cite{kim2014convolutional} as a base model. To examine the robustness of the proposed approach, we intentionally flip the class labels with $0\%$ to $70\%$ label noise, in other words: ~$p ~\in ~\{0.0, 0.1, 0.2, 0.3, 0.4, 0.5, 0.6, 0.7\}$, and observe the effect of different types of label flipping, such as uniform (\emph{Uni}) and random (\emph{Rand}) label flipping, along with instance-dependent label noise. For all the experiments, we use early stopping based on validation set accuracy where the class labels in validation are also corrupted.

We indicate the performance of a standard deep network \emph{Without Noise model (WoNM)} on the noisy label dataset. We also plot the results for the stacked \emph{Noise Model Without Regularization (NMWoRegu)} and stacked \emph{Noise Model With Regularization (NMwRegu)}. Unless otherwise stated, in all the deep networks with the stacked noise model, we initialize the noise layer parameters as an identity matrix. We further analyze the effect of the noise layer initialization on the overall performance. We define \emph{TDwRegu} as the stacked noise model with regularization, initialized with true injected noise distribution and \emph{RandwRegu} as the stacked noise model with regularization, initialized randomly. We run all experiments five times and report the mean accuracy.

\subsection{Datasets}


\begin{table}
\resizebox{\columnwidth}{!}{\begin{tabular}{c|l||c|c|c|c|c} \hline
\multirow{5}{*}{\rotatebox[origin=c]{90}{\parbox[c]{2cm}{\centering Text Data}}} &Dataset&K&L&N & T& Type\\ \cline{2-7}
&SST-2 & 2& 19& 76961 & 1821& Balanced\\
&Trec & 6& 10& 5000 & 500& Not Balanced\\
&AG-News & 4& & 110K & 10K& Balanced\\
&DbPedia & 14& 29& 504K & 70K& Balanced\\ 
\specialrule{1.5pt}{1pt}{1pt}
\end{tabular}}
\vspace{0.05in}
\caption{Summary of text classification datasets; K: denotes the number of classes, L: represents the average length of sentence, N: denotes the number of training samples, T: represents the number of test samples, Type: describes whether the dataset is balanced.}
\label{table:text_datasets}
\end{table}

\begin{table*}[!t]
\resizebox{\textwidth}{!}{\begin{tabular}{c|l||c||ccccccc||cccccccc} \hline \hline
 \multirow{8}{*}{\rotatebox[origin=c]{90}{\parbox[c]{2cm}{\centering SST-2}}} &  &Batch Size
&\multicolumn{7}{c||}{\textbf{50}}&\multicolumn{8}{c}{\textbf{100}}\\ \cline{3-18}
 &     &Label Flips              & \multicolumn{7}{c||}{\underline{Random}}                          & \multicolumn{8}{c}{\underline{Random}}                           \\ \cline{2-18}
&Noise\% & Clean Data& 10  & 20 & 30 & 40 & 45 & 47 & 50                & 0 & 10& 20& 30 &40 &45&47 &50\\ \cline{2-18}
&WoNM          & $87.27\%$&$83.29\%$&$79.08\%$&$73.42\%$&$64.03\%$&$58.1\%$& $54.73\%$&$49.7\%$      & $86.53\%$&$81.44\%$&$75.58\%$&$71.88\%$&$63.39\%$&$57.12\%$& $55.81\%$&$52.32\%$ \\ \cline{2-18}

&TDwRegu01         &$86.88\%$&$85.37\%$&$  84.92\%$&$\textbf{83.29\%} $&$\textbf{78.53\%} $&$\textbf{74.01\%}$&$51.95\%$&$\textbf{49.5\%}$         & $\textbf{86.88\%}$&$\textbf{84.88\%}$&$\textbf{85.08\%}$&$\textbf{82.41\%}$&$\textbf{76.09\%}$&$\textbf{70.10\%}$& $\textbf{58.98\%}$&$49.86\%$ \\ 


&NMWoRegu               &$\textbf{87.28\%}$& $\textbf{86.2\%}$&$84.07\%$&$81.29\% $&$ 70.42\%$&$62.27\% $&$\textbf{55.76\%}$&$ 48.42\%$     & $86.66\%$&$84.72\%$&$83.03\%$&$78.2\%$&$66.65\%$&$61.32\%$& $57.11\%$&$\textbf{52.24\%}$ \\ \cline{2-18}

&NMwRegu001        &$86.08\% $&$85.01\%$&$ 83.82\% $&$81.97\%$&$73.18\%$&$62.18\%$&$55.63\%$&$48.87\% $        & $\textbf{86.51\%}$&$\textbf{85.26\%}$&$84.37\%$&$81.05\%$&$69.54\%$&$60.89\%$& $56.8\%$&$51.6\%$ \\

&NMwRegu01             &$\textbf{87.78\%}$&$\textbf{86.04\%}$&$  \textbf{85.04\%}$&$\textbf{82.7\%}$&$\textbf{77.43\%}$&$\textbf{66.96\%}$&$\textbf{61.5\%}$&$\textbf{49.08\%}$    & $86.33\%$&$85.17\%$&$\textbf{85.10\%}$&$\textbf{81.9\%}$&$\textbf{76.2\%}$&$\textbf{65.47\%}$& $\textbf{58.92\%}$&$\textbf{52.46\%}$  \\ \hline
\specialrule{1.5pt}{1pt}{1pt}
\multirow{16}{*}{\rotatebox[origin=c]{90}{\parbox[c]{2cm}{\centering Trec}}} & &Batch Size
&\multicolumn{15}{c}{\textbf{10}} \\ \cline{3-18}
&     &Label Flips              & \multicolumn{7}{c||}{\underline{Uniform}}                          & \multicolumn{8}{c}{\underline{Random}}                           \\ \cline{2-18}
&Noise\% & Clean data& 10  & 20 & 30 & 40 & 50 & 60 & 70                & 0 & 10& 20& 30 &40 &50&60 &70\\ \cline{2-18} 
&WoNM          	   &$92.8\%$&$87.6\%$&$83.6\% $&$75.87\%$&$67.27\%$&$57.4\% $& $46.27\%$&$42.8\% $      &$92.8\%$&$85.93\%$&$82.2\%$&$74.0\%$&$68.4\%$&$53.53\%$&$48.2\%$&$31.47\%$\\ \cline{2-18}

&TDwRegu01         &$50.87\%$&$45.33\%$&$  45.4\%$&$36.33\%$&$25.87\%$&$28.33\%$&$16.87\%$&$16.87\%$         &$50.87\%$&$56.4\%$&$36.8\%$&$24.0\%$&$25.47\%$&$22.6\%$&$18.8\%$&$22.6\% $\\ 


&NMWoRegu          &$92.33\%$& $88.07\%$&$84.67\%$&$76.4 \%$&$ 68.47\%$&$58.4\%$&$50.07\%$&$ 41.33\%$     &$92.07\%$&$85.87\%$&$84.27\%$&$72.47\%$&$66.53\%$&$50.13\%$&$44.6\%$&$33.0\%$\\ \cline{2-18}

&NMwRegu001        &$92.47\%$&$90.53\%$&$ 88.07\%$&$81.6\%$&$73.47\%$&$64.07\%$&$55.87\%$&$43.67\% $        &$92.4\%$&$88.53\%$&$86.4\%$&$77.2\%$&$67.67\%$&$54.67\%$&$47.93\%$&$\textbf{34.87\%}$\\

&NMwRegu01         &$\textbf{92.73\%}$&$\textbf{90.8\%}$&$  \textbf{89.53\%}$&$\textbf{88.67\%}$&$\textbf{84.93\%}$&$\textbf{79.67\%}$&$\textbf{69.67\%}$&$\textbf{52.4\% }$     &$\textbf{92.7\%}$&$\textbf{90.33\%}$&$\textbf{90.6\%}$&$\textbf{86.47\%}$&$\textbf{83.07\%}$&$\textbf{70.93\%}$&$\textbf{65.2\%}$&$33.4\%$\\ \cmidrule[2pt]{2-18}

& &Batch Size
&\multicolumn{15}{c}{\textbf{50}} \\ \cline{3-18}
&     &Label Flips              & \multicolumn{7}{c||}{\underline{Uniform}}                          & \multicolumn{8}{c}{\underline{Random}}                           \\ \cline{2-18}
&Noise\% & 0& 10  & 20 & 30 & 40 & 50 & 60 & 70                & 0 & 10& 20& 30 &40 &50&60 &70\\ \cline{2-18} 
&WoNM          	   &$92.8\%$&$87.27\%$&$83.07\% $&$75.00\%$&$69.13\%$&$61.53\% $& $50.13\%$&$39.8\% $      &$92.8\%$&$86.00\%$&$81.2\%$&$76.2\%$&$64.07\%$&$52.4\%$&$47.4\%$&$34.13\%$\\ \cline{2-18}

&TDwRegu01         &$55.73\%$&$50.4\%$&$  44.73\%$&$39.6\%$&$22.27\%$&$25.67\%$&$14.93\%$&$21.00\%$         &$55.73\%$&$45\%$&$44.93\%$&$27.73\%$&$27.87\%$&$22.6\%$&$17.87\%$&$22.6\% $\\ 


&NMWoRegu          &$92.6\%$& $87.73\%$&$83.33\%$&$76.33 \%$&$ 70.67\%$&$56.8\%$&$48.2\%$&$ 39.67\%$     &$92.60\%$&$85.27\%$&$83.00\%$&$73.6\%$&$65.8\%$&$50.4\%$&$45.93\%$&$30.73\%$\\ \cline{2-18}

&NMwRegu001        &$92.53\%$&$90.73\%$&$ 87.20\%$&$82.53\%$&$73.93\%$&$65.07\%$&$52.87\%$&$44.60\% $        &$92.53\%$&$88.\%$&$87.2\%$&$79.07\%$&$71.2\%$&$51.67\%$&$49.00\%$&$33.40\%$\\

&NMwRegu01         &$\textbf{92.53\%}$&$\textbf{91.33\%}$&$  \textbf{90.27\%}$&$\textbf{88.47\%}$&$\textbf{83.87\%}$&$\textbf{77.87\%}$&$\textbf{68.73\%}$&$\textbf{55.67\% }$     &$\textbf{92.53\%}$&$\textbf{90.00\%}$&$\textbf{90.2\%}$&$\textbf{85.93\%}$&$\textbf{82.6\%}$&$\textbf{71.4\%}$&$\textbf{67.33\%}$&$\textbf{37.53\%}$\\\hline

\specialrule{1.5pt}{1pt}{1pt}
\multirow{16}{*}{\rotatebox[origin=c]{90}{\parbox[c]{2cm}{\centering AG-News}}} & &Batch Size
&\multicolumn{15}{c}{\textbf{100}} \\ \cline{3-18}
&     &Label Flips              & \multicolumn{7}{c||}{\underline{Uniform}}                          & \multicolumn{8}{c}{\underline{Random}}                           \\ \cline{2-18}
&Noise\% & 0& 10  & 20 & 30 & 40 & 50 & 60 & 70                & 0 & 10& 20& 30 &40 &50&60 &70\\ \cline{2-18} 
&WoNM          	   &$92.31\%$&$89.96\%$&$87.42\% $&$84.55\%$&$79.96\%$&$75.42\% $& $68.78\%$&$59.94\% $      &$92.31\%$&$89.71\%$&$86.11\%$&$79.05\%$&$76.04\%$&$65.09\%$&$45.79\%$&$38.12\%$\\ \cline{2-18}

&TDwRegu01         &$92.47\%$&$92.25\%$&$92.15\%$&$92.04\%$&$84.87\%$&$77.56\%$&$63.13\%$&$47.83\%$         &$92.68\%$&$92.09\%$&$91.99\%$&$61.81\%$&$62.44\%$&$70.26\%$&$24.99\%$&$38.12\% $\\ 


&NMWoRegu          &$91.94\%$& $91.89\%$&$91.21\%$&$90.51 \%$&$ 89.29\%$&$88.02\%$&$86.25\%$&$ 79.88\%$     &$91.97\%$&$91.79\%$&$91.00\%$&$90.04\%$&$88.82\%$&$86.49\%$&$\textbf{77.66\%}$&$43.01\%$\\ \cline{2-18}

&NMwRegu001        &$92.47\%$&$92.21\%$&$ 91.82\%$&$91.21\%$&$90.71\%$&$89.61\%$&$88.43\%$&$85.32\% $        &$\textbf{92.62\%}$&$92.14\%$&$91.5\%$&$91.07\%$&$90.2\%$&$88.68\%$&$64.01\%$&$55.11\%$\\

&NMwRegu01         &$\textbf{92.55\%}$&$\textbf{92.23\%}$&$  \textbf{92.2\%}$&$\textbf{91.98\%}$&$\textbf{91.7\%}$&$\textbf{91.23\%}$&$\textbf{90.54\%}$&$\textbf{89.78\%} $     &$92.57\%$&$\textbf{92.23\%}$&$\textbf{91.96\%}$&$\textbf{91.69\%}$&$\textbf{91.13\%}$&$\textbf{90.77\%}$&$76.64\%$&$\textbf{62.04\%}$\\\cmidrule[2pt]{2-18}

& &Batch Size
&\multicolumn{15}{c}{\textbf{1024}} \\ \cline{3-18}
&     &Label Flips              & \multicolumn{7}{c||}{\underline{Uniform}}                          & \multicolumn{8}{c}{\underline{Random}}                           \\ \cline{2-18}
&Noise\% & 0& 10  & 20 & 30 & 40 & 50 & 60 & 70                & 0 & 10& 20& 30 &40 &50&60 &70\\ \cline{2-18} 
&WoNM          	   &$92.42\%$&$89.77\%$&$87.04\% $&$84.07\%$&$79.77\%$&$74.54\% $& $67.59\%$&$59.41\% $      &$92.29\%$&$89.47\%$&$85.78\%$&$80.51\%$&$75.99\%$&$65.55\%$&$45.50\%$&$39.75\%$\\ \cline{2-18}

&TDwRegu01         &$92.61\%$&$92.37\%$&$92.18\%$&$92.07\%$&$84.92\%$&$62.74\%$&$63.43\%$&$47.59\%$         &$92.54\%$&$92.34\%$&$91.82\%$&$53.81\%$&$69.04\%$&$48.88\%$&$25.05\%$&$46.9\% $\\ 


&NMWoRegu          &$92.16\%$& $91.51\%$&$90.80\%$&$89.58 \%$&$ 85.58\%$&$79.96\%$&$70.79\%$&$62.89\%$     &$92.22\%$&$91.61\%$&$90.33\%$&$86.92\%$&$82.61\%$&$71.49\%$&$48.96\%$&$39.96\%$\\ \cline{2-18}

&NMwRegu001        &$92.4\%$&$92.13\%$&$ 91.88\%$&$91.46\%$&$90.14\%$&$89.07\%$&$86.96\%$&$80.94\% $        &$92.54\%$&$91.87\%$&$91.38\%$&$90.42\%$&$99.18\%$&$86.78\%$&$75.74\%$&$50.11\%$\\

&NMwRegu01         &$\textbf{92.66\%}$&$\textbf{92.2\%}$&$  \textbf{92.29\%}$&$\textbf{92.09\%}$&$\textbf{91.7\%}$&$\textbf{91.24\%}$&$\textbf{90.72\%}$&$\textbf{89.88\%} $     &$\textbf{92.57\%}$&$\textbf{92.11\%}$&$\textbf{91.99\%}$&$\textbf{91.57\%}$&$\textbf{91.2\%}$&$\textbf{90.5\%}$&$\textbf{77.93\%}$&$\textbf{61.12\%}$\\\hline

\specialrule{1.5pt}{1pt}{1pt}
\multirow{15}{*}{\rotatebox[origin=c]{90}{\parbox[c]{2cm}{\centering DBpedia}}} 
& &Batch Size
&\multicolumn{15}{c}{\textbf{512}} \\ \cline{3-18}
&     &Label Flips              & \multicolumn{7}{c||}{\underline{Uniform}}                          & \multicolumn{8}{c}{\underline{Random}}                           \\ \cline{2-18}
&Noise\% & Clean data& 30  & 50 & 70 & 75 & 80 & 85 & 90                & 0 & 30& 50& 70 &75 & 80 & 85 & 90\\ \cline{2-18} 
&WoNM          	   &$99.01\%$&$95.19\%$&$89.59\% $&$74.01\%$&$67.73\%$&$57.87\%$& $47.48\%$&$34.01\%$      &$99.01\%$&$94.72\%$&$86.08\%$&$62.87\%$&$53.13\%$&$40.78\%$&$26.6\%$&$12.42\%$\\ \cline{2-18}



&NMWoRegu          &$98.93\%$& $95.07\%$&$90.2\%$&$78.32\%$&$ 73.65\%$&$66.24\%$&$54.24\%$&$40.9\%$     &$98.93\%$&$93.55\%$&$84.53\%$&$25.96\%$&$54.84\%$&$42.96\%$&$29.25\%$&$12.97\%$\\ \cline{2-18}

&NMwRegu001        &$\textbf{99.04\%}$&$\textbf{98.94\%}$&$ \textbf{98.81\%}$&$\textbf{98.61\%}$&$\textbf{98.52\%}$&$\textbf{98.33\%}$&$\textbf{98.13\%}$&$\textbf{97.53\%}$        &$\textbf{99.04\%}$&${98.93\%}$&${98.82\%}$&${98.62\%}$&$\textbf{98.48\%}$&$\textbf{98.33\%}$&$\textbf{89.00\%}$&$11.36\%$\\

&NMwRegu01         &$99.01\%$&$98.89\%$&$  98.71\%$&$98.45\%$&$98.32\%$&$98.10\%$&$97.76\%$&$97.15\%$     &$98.92\%$&$\textbf{99.01\%}$&$\textbf{98.88\%}$&$\textbf{98.72\%}$&$98.10\%$&$97.67\%$&$38.62\%$&$\textbf{16.27\%}$\\ \cmidrule[2pt]{2-18}

& &Batch Size
&\multicolumn{15}{c}{\textbf{1024}} \\ \cline{3-18}
&     &Label Flips              & \multicolumn{7}{c||}{\underline{Uniform}}                          & \multicolumn{8}{c}{\underline{Random}}                           \\ \cline{2-18}
&Noise\% & Clean data& 30  & 50 & 70 & 75 & 80 & 85 & 90                & 0 & 30& 50& 70 &75 & 80 & 85 & 90\\ \cline{2-18} 
&WoNM          	   &$98.96\%$&$97.93\%$&$96.47\% $&$90.49\%$&$68.07\%$&$59.78\%$& $48.06\%$&$55.29\%$      &$98.96\%$&$94.75\%$&$86.36\%$&$63.75\%$&$53.39\%$&$40.87\%$&$26.18\%$&$11.9\%$\\ \cline{2-18}



&NMWoRegu          &$98.87\%$& $97.37\%$&$95.71\%$&$89.54\%$&$72.79\%$&$66.49\%$&$55.27$&$60.7\%$     &$98.87\%$&$93.96\%$&$85.6\%$&$44.85\%$&$54.32\%$&$42.21\%$&$28.63\%$&$12.51\%$\\ \cline{2-18}

&NMwRegu001        &$\textbf{98.97\%}$&$\textbf{98.9\%}$&$ \textbf{98.79\%}$&$\textbf{98.53\%}$&$\textbf{98.50\%}$&$\textbf{98.32\%}$&$\textbf{98.19\%}$&$\textbf{97.27\%}$        &$\textbf{98.97\%}$&$98.83\%$&$98.51\%$&$98.1\%$&$\textbf{98.49\%}$&$\textbf{98.32\%}$&$\textbf{83.79\%}$&$10.51\%$\\

&NMwRegu01         &$98.92\%$&$98.79\%$&$  98.58\%$&$98.26\%$&$98.32\%$&$98.09\%$&$97.79\%$&$96.54\%$     &$98.92\%$&$\textbf{98.88\%}$&$\textbf{98.72\%}$&$\textbf{98.35\%}$&$98.12\%$&$97.72\%$&$33.10\%$&$\textbf{15.94\%}$\\\hline
\specialrule{1.5pt}{1pt}{1pt}
\end{tabular}}
\caption{Test performance for different text classification datasets}
\label{table:result_text}
\end{table*}

Here, we describe all the text classification datasets used to evaluate the performance of the proposed approach. The base model architecture is the same for all datasets. For each set, we tune the number of filter windows and filter lengths using the development set. Along with the description, we also provide the hyper-parameters we selected for each. Table \ref{table:text_datasets} summarizes the basic statistic of the datasets.

\begin{enumerate}
	\item SST-2\footnote{\url{http://nlp.stanford.edu/sentiment/}} \cite{socher2011semi}: Stanford Sentiment Treebank dataset for predicting the sentiment of movie reviews. The classification task involves detecting positive or negative reviews. Using the base model with clean labels we obtain classification accuracy of $87.27\%$. For this dataset, the base model network architecture consists of an input and embedding layer + $[3,4,5]$ feature windows with 100 feature maps each and dropout rate $0.5$ with batch size 50.

	\item TREC\footnote{\url{http://cogcomp.cs.illinois.edu/Data/QA/QC/}} \cite{voorhees1999trec}: A question classification dataset consisting of fact based questions divided into broad semantic categories. We use a six-class version of TREC dataset. For this dataset, the base model network architecture consists of an input and embedding layer + $[3]$ one feature windows with 100 feature maps and dropout rate $0.5$ with batch size 10.

	\item Ag-News\footnote{\label{foot:ag}\url{http://www.di.unipi.it/~gulli/AG_corpus_of_news_articles.html}} \cite{zhang2015character}: A large-scale, four-class topic classification dataset. It contains approx 110K training samples. For this dataset, the base model network architecture consists of Input layer + Embedding layer + $[3, 4, 5]$ feature windows with 200 feature maps and dropout rate $0.5$ with batch size 100.

	\item DBpedia\footref{foot:ag} \cite{zhang2015character}: A large scale 14-class topic classification dataset containing $36K$ training samples per category. For this dataset, the base model network architecture consists of Input layer + Embedding layer + $[1,2,3,4,5,7]$ feature windows with 400 feature maps each and dropout rate $0.5$ with batch size 1024.
\end{enumerate}

For all the datasets, we use Rectified Linear Units (ReLU) and fix the base model architecture. We use early stopping on dev sets for all the datasets. We run all the experiments 5 times and report the average classification accuracy in Table \ref{table:result_text}. We train all the networks end-to-end via stochastic gradient descent over shuffled mini-batches with the Adadelta update rule \cite{zeiler2012adadelta} except for the DBpedia, where we use SGD. In order to improve base model performance, we initialize the word embedding layer with the publicly available \emph{word2vec} word vectors \cite{mikolov2013distributed} for all the datasets except for DBpedia, where we use \emph{GloVe} embeddings \cite{pennington2014glove}.


\section{Results and Discussion}
\label{sec:results}
We evaluate the performance of our model in Table \ref{table:result_text} for each datasets in the presence of uniform and random label noise and compare the performance with the base model (\emph{WoNM}) as our baseline. For the other datasets, the proposed approach is significantly better than the baseline for both types of label noise. For all datasets, we observe a gain of approximately $30\%$ w.r.t baseline in the presence of extreme label noise. Interestingly, if we assume an oracle to determine prior knowledge of true noise distribution (\emph{TDwRegu01}), it does not necessarily improve classification performance, especially for multi-class classification problems. For binary classification, using the SST-2 dataset, we did observe that the noise model initialized with the true noise distribution works better than all the other models. 

\subsection{Effect of different regularizers} 
\begin{figure}[!ht]
\centering
	\begin{subfigure}{0.31\columnwidth}
		\includegraphics[width=\columnwidth]{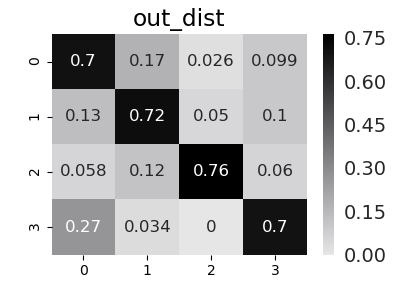}
		\caption{}
		\label{fig::AG_input}
	\end{subfigure}
	~
	\begin{subfigure}{0.31\columnwidth}
		\includegraphics[width=\columnwidth]{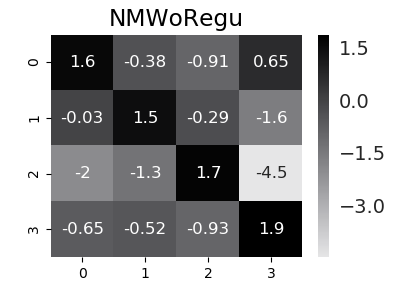}
		\caption{}
		\label{fig::AG_noregu}
	\end{subfigure}
	~
	\begin{subfigure}{0.31\columnwidth}
		\includegraphics[width=\columnwidth]{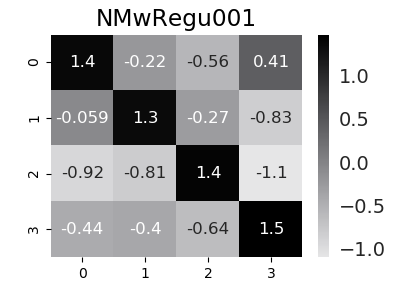}
		\caption{}
		\label{fig::AG_regu001}
	\end{subfigure}
	
	\begin{subfigure}{0.31\columnwidth}
		\includegraphics[width=\columnwidth]{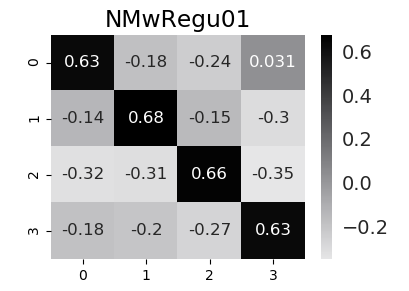}
		\caption{}
		\label{fig::AG_regu01}
	\end{subfigure}
	~
	\begin{subfigure}{0.31\columnwidth}
		\includegraphics[width=\columnwidth]{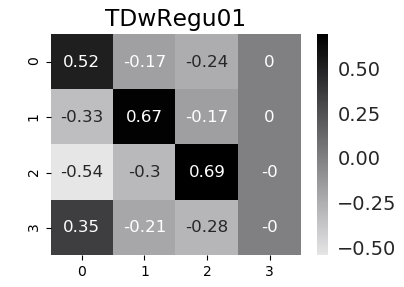}
		\caption{}
		\label{fig::AG_TDregu01}
	\end{subfigure}
	~
	\begin{subfigure}{0.31\columnwidth}
		\includegraphics[width=\columnwidth]{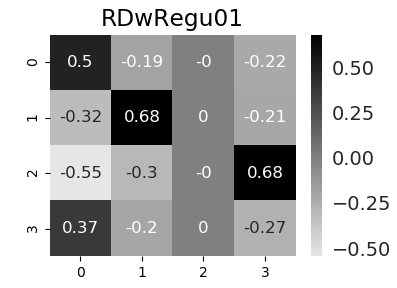}
		\caption{}
		\label{fig::AG_RDregu01}
	\end{subfigure}
	\caption{AG-News Dataset: a) Input random label noise; (b-f) learned weight matrix learned by different noise models.}
	\label{fig:AG_news_weight}
	\vspace{-0.1in}
\end{figure}

The \emph{NMwRegu01} performs better in all cases for both types of label noise. We plot the weight matrix learned by all the noise models in all the noise regimes. For brevity, we only plot the weight matrix for AG-News datasets with $30\%$ label noise in Fig.\ref{fig:AG_news_weight}. We find that $l_2$ regularization diffuses the diagonal weight elements and learns more smoothed off-diagonal elements which resemble the corresponding input label noise distribution in Fig. \ref{fig::AG_regu01}. This also means that, without regularization, the noise model has less ability to diffuse the diagonal elements which leads to poor classification performance. Therefore, we use a regularizer ($l_2$) to diffuse the diagonal entries. 

In some cases, especially for low label noise, we find the $l_2$ regularization with a small penalty works better than a large penalty since, for low label noise, learning a less diffuse noise is beneficial. The proposed approach scales to a large number of label categories, as evident from the experiments on DBpedia dataset in the last row of Table .\ref{table:result_text}.


\subsection{Effect of different scaling factors on noise layer initialization} 
\begin{figure}[!ht]
\centering
	\begin{subfigure}{0.45\columnwidth}
		\includegraphics[width=\columnwidth]{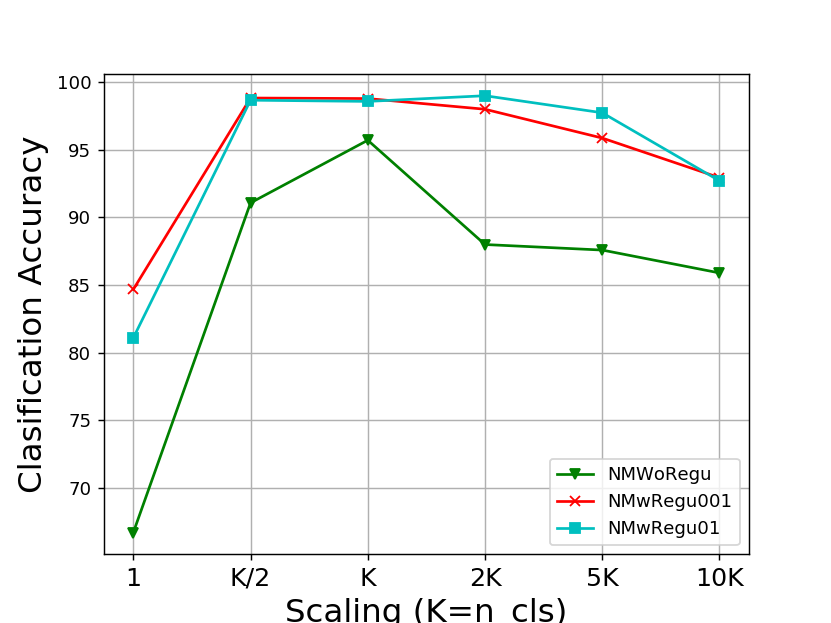}
		\caption{Classification accuracy}
		\label{fig::dbpedia_scaling}
	\end{subfigure}
	~
	\begin{subfigure}{0.45\columnwidth}
		\includegraphics[width=\columnwidth]{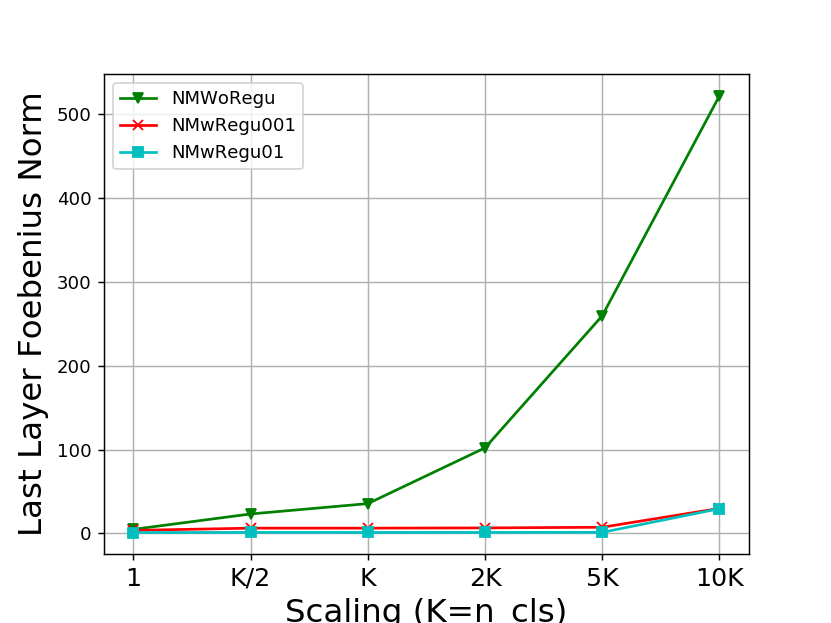}
		\caption{Noise model norm}
		\label{fig::dbpedia_norm}
	\end{subfigure}
	\caption{Effect of noise model initialization scaling on the classification performance}
	\label{fig:dbpedia_sc}
\end{figure}

We initialize the noise model weights as identity matrices with gain equal to the number of classes (gain = $K$) for all experiments. We observe the effect of different gain values on the overall performance of the proposed network in Fig. \ref{fig:dbpedia_sc}. We plot the classification performance for the  DBpedia dataset with $50\%$ random noise. For each noise model in Fig. \ref{fig::dbpedia_scaling}, we find that setting the gain to $K$ works best and any other gain results in poor performance.

In Fig. \ref{fig::dbpedia_norm} we plot the Frobenius norm of the learned noise model weights with respect to the different gain values. We find that, using the high gain initialization, the model learns a high noise model norm, resulting in poor classification performance. This finding provides support to the claim in \cite{liao2018surprising} that ``higher capacity leads to high test errors.''

\subsection{Effect of Batch size} 
\begin{figure}[!ht]
\centering
	\begin{subfigure}{0.45\columnwidth}
		\includegraphics[width=\columnwidth]{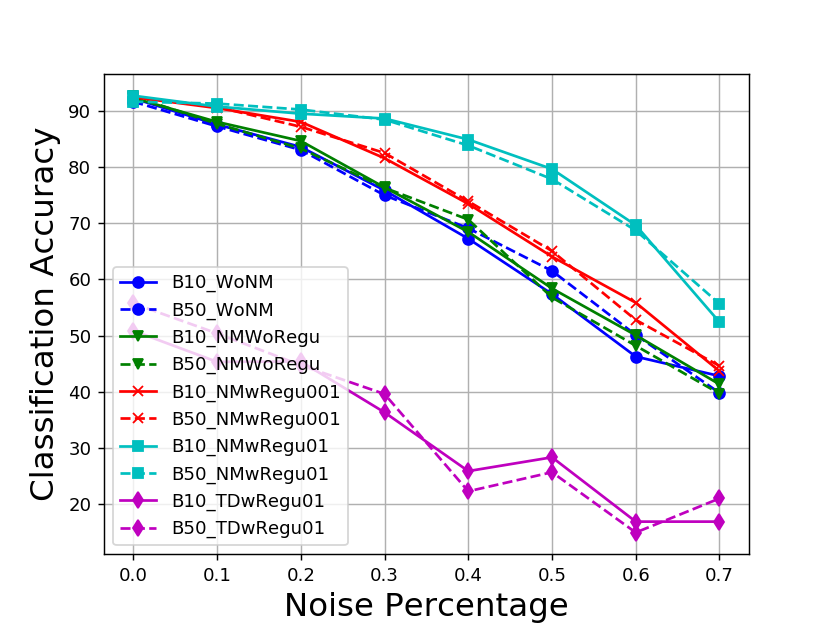}
		\caption{Trec [Uniform]}
		\label{fig:Utrec_batchsize}
	\end{subfigure}
	~
	\begin{subfigure}{0.45\columnwidth}
		\includegraphics[width=\columnwidth]{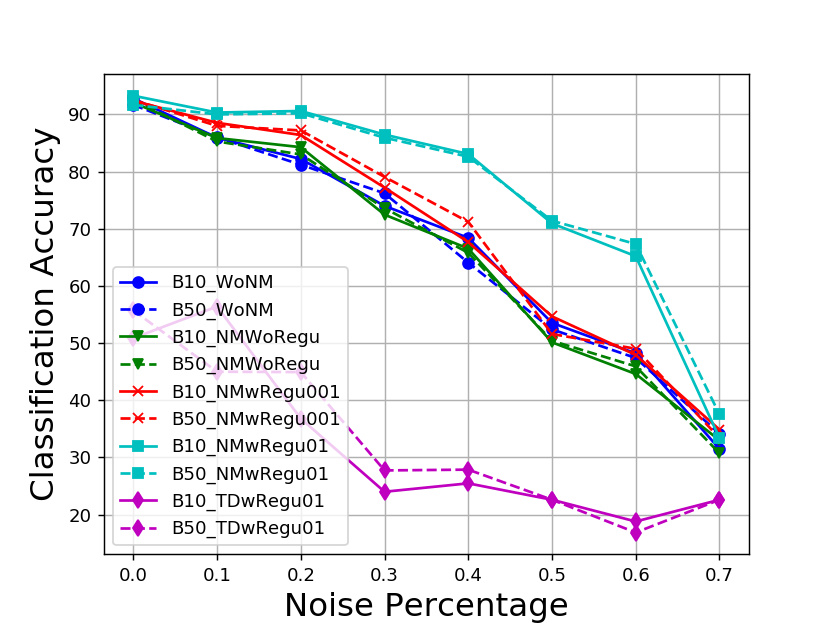}
		\caption{Trec [Random]}
		\label{fig:Rtrec_batchsize}
	\end{subfigure}

		\begin{subfigure}{0.45\columnwidth}
		\includegraphics[width=\columnwidth]{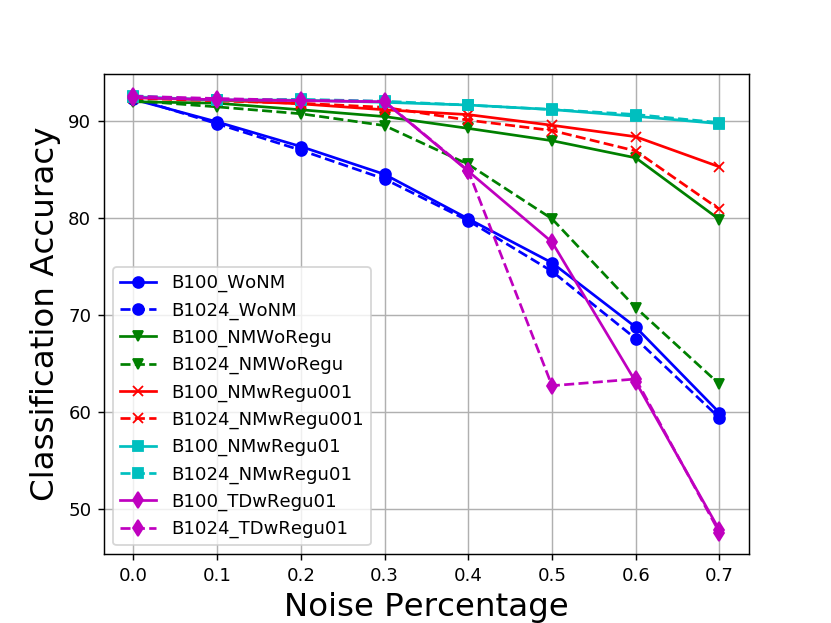}
		\caption{AG-News [Uniform]}
		\label{fig:UAG_batchsize}
	\end{subfigure}
	~
	\begin{subfigure}{0.45\columnwidth}
		\includegraphics[width=\columnwidth]{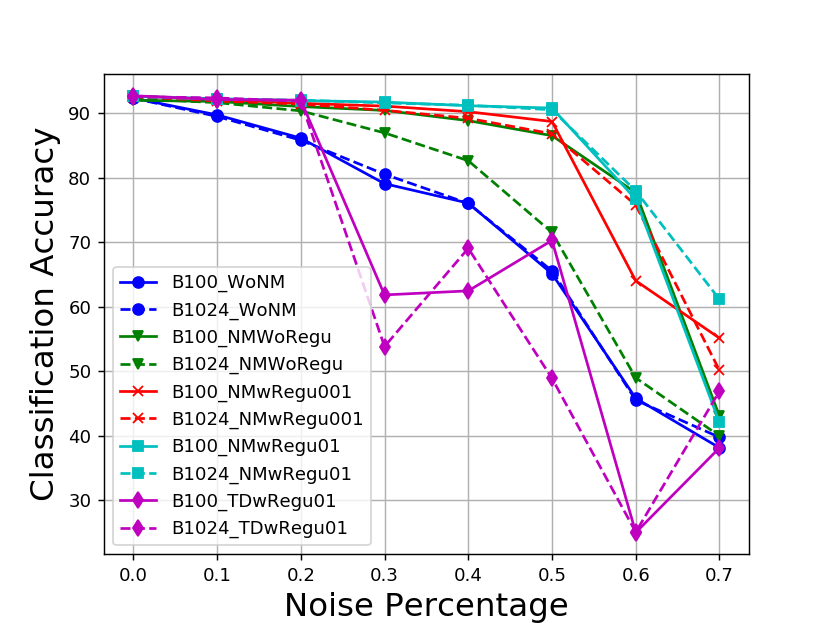}
		\caption{AG-News [Random]}
		\label{fig:RAG_batchsize}
	\end{subfigure}
	
	\begin{subfigure}{0.45\columnwidth}
		\includegraphics[width=\columnwidth]{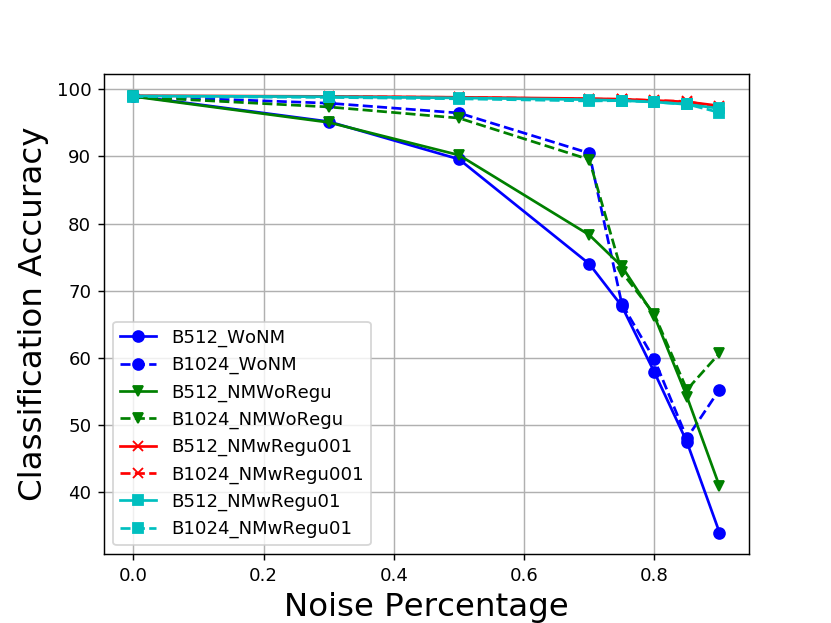}
		\caption{DBpedia [Uniform]}
		\label{fig:Udbpedia_batchsize}
	\end{subfigure}
	~
	\begin{subfigure}{0.45\columnwidth}
		\includegraphics[width=\columnwidth]{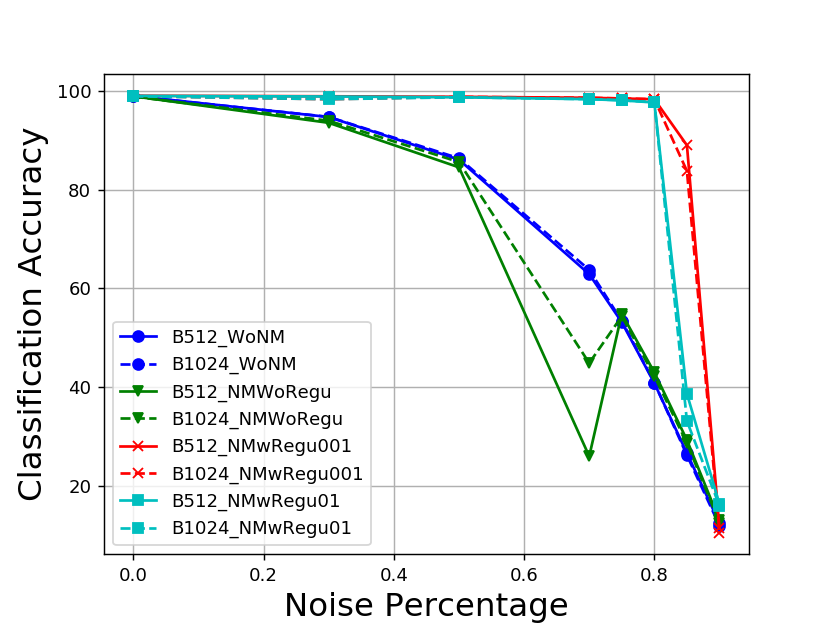}
		\caption{DBpedia [Random]}
		\label{fig:Rdbpedia_batchsize}
	\end{subfigure}
	\caption{Effect of batch size on label noise classification for different datasets}
	\label{fig::batchsize}
\end{figure}

We also observe the effect of different batch sizes on performance as described in \cite{rolnick2017deep}. For all datasets, we do observe small performance gains for highly non-uniform noisy labels, for instance $70\%$, in Fig. \ref{fig::batchsize} column 2. However, for uniform label flips, we do not observe performance gains with increasing batch size. 

\subsection{Instance Dependent label noise}
\begin{figure}[!ht]
\centering
	\begin{subfigure}{0.31\columnwidth}
		\includegraphics[width=\columnwidth]{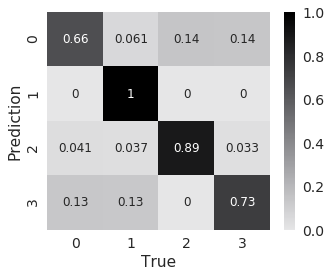}
		\caption{}
		\label{fig::AG_CC_input}
	\end{subfigure}
	~
	\begin{subfigure}{0.31\columnwidth}
		\includegraphics[width=\columnwidth]{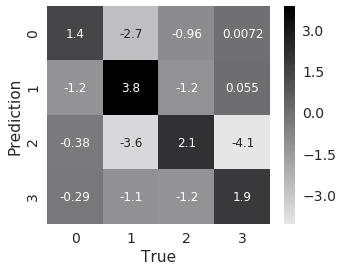}
		\caption{}
		\label{fig::AG_CC_actua;}
	\end{subfigure}
	~
	\begin{subfigure}{0.31\columnwidth}
		\includegraphics[width=\columnwidth]{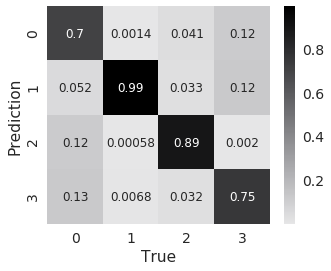}
		\caption{}
		\label{fig::AG_CC_norm}
	\end{subfigure}
	\caption{AG-News Dataset: a) input instance dependent label noise; b) learned weight matrix by proposed approach; c) column normalization of (b).}
	\label{fig:AG_CC}
\end{figure}

We further investigate the performance of the proposed approach on instance-dependent label noise by flipping each class labels with different noise percentages as shown in Fig. \ref{fig::AG_CC_input}. For brevity, we present results on AG-News dataset in Fig. \ref{fig:AG_CC}. On this type of label noise, the performance of proposed approach is far better than the baseline with a performance improvement of $\sim 6\%$. The learned noise model by the proposed approach is shown in Fig. \ref{fig::AG_CC_actua;} and we show the normalized weight matrix in Fig. \ref{fig::AG_CC_norm}. We observe that the learned noise model is able to capture the input label noise statistics and is highly correlated to the input noise distribution with Pearson Correlation Coefficient $0.988$.

\subsection{Understanding Noise Model}

\begin{table}
\resizebox{\columnwidth}{!}{\begin{tabular}{l|ccc||ccc}\hline
\multicolumn{1}{c|}{ }&\multicolumn{3}{c||}{ TRB}&\multicolumn{3}{c}{TRPr}\\ \hline
Data(N$\%$)&WoNM&Noisy&True&NMwRegu01&Noisy&True\\ \hline
SST2 (40\%)&70.24&70.95&79.24&82.32&73.90&83.25\\ \hline
AG (70\%)&59.70&52.44&79.18&90.33&86.27&89.4\\ \hline
AG (60\%)&83.25&68.8&88.28&90.45&87.77&90.78\\ \hline
Trec (40\%)&66.80&63.4&79.0&73.40&69.6&83.2\\ \hline
Trec (20\%)&83.6&80.0&86.0&87.40&83.6&90.0\\ \hline
\end{tabular}}
\caption{SVM Classification}
\label{table::svm_performance}
\end{table}

\begin{figure*}[!ht]
\centering
\begin{subfigure}{\textwidth}
	\begin{subfigure}{0.239\columnwidth}
		\includegraphics[width=\columnwidth]{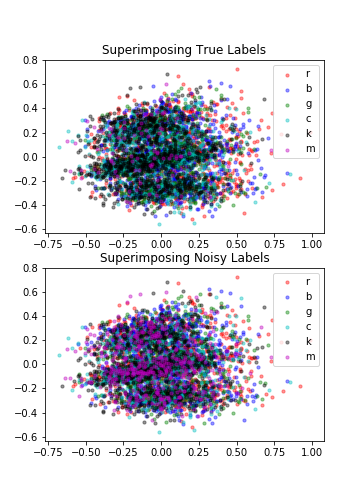}
		\caption*{Iteration 0}
	\end{subfigure}
	~
	\begin{subfigure}{0.239\columnwidth}
		\includegraphics[width=\columnwidth]{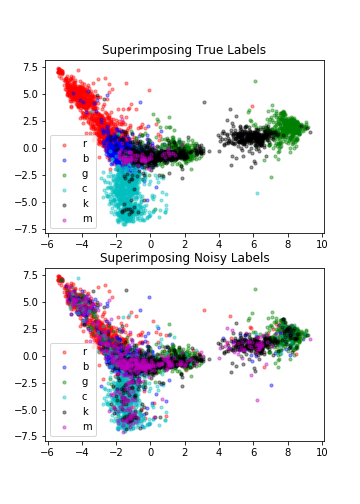}
		\caption*{Iteration 5}
	\end{subfigure}
	~
	\begin{subfigure}{0.239\columnwidth}
		\includegraphics[width=\columnwidth]{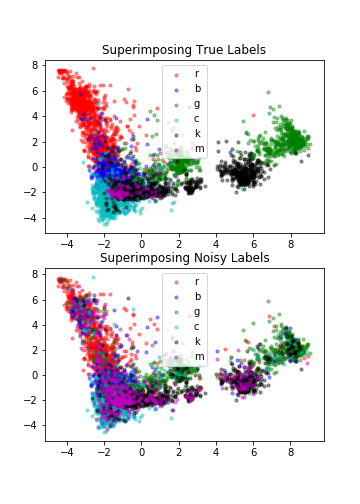}
		\caption*{Iteration 10}
	\end{subfigure}
	~
	\begin{subfigure}{0.239\columnwidth}
		\includegraphics[width=\columnwidth]{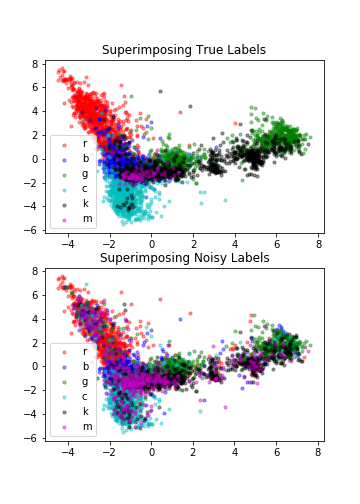}
		\caption*{Iteration 18}
	\end{subfigure}
	\caption{Proposed model}
	\label{fig::tsne_Nmwregu}
\end{subfigure}

\begin{subfigure}{\textwidth}
	\begin{subfigure}{0.239\columnwidth}
		\includegraphics[width=\columnwidth]{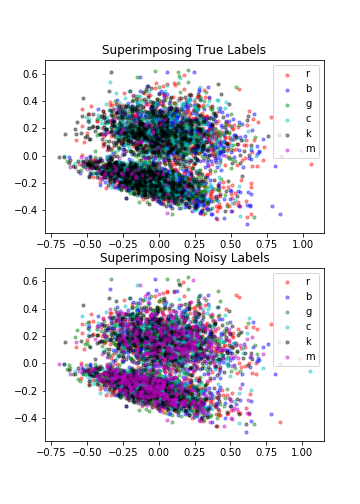}
		\caption*{Iteration 0}
	\end{subfigure}
	~
	\begin{subfigure}{0.239\columnwidth}
		\includegraphics[width=\columnwidth]{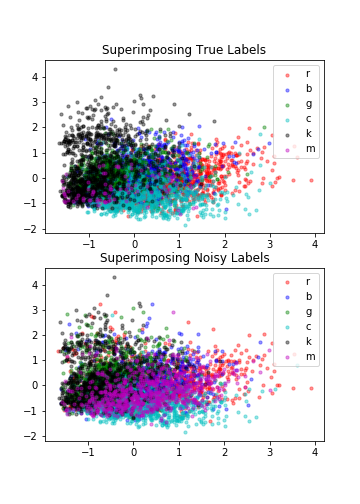}
		\caption*{Iteration 5}
	\end{subfigure}
	~
	\begin{subfigure}{0.239\columnwidth}
		\includegraphics[width=\columnwidth]{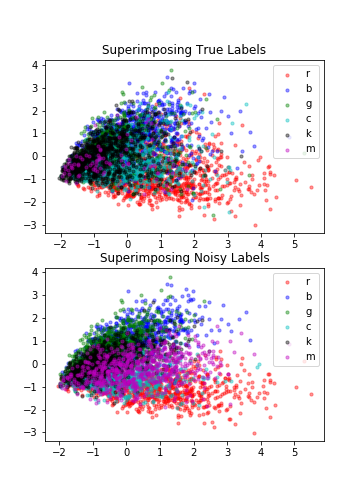}
		\caption*{Iteration 10}
	\end{subfigure}
	~
	\begin{subfigure}{0.239\columnwidth}
		\includegraphics[width=\columnwidth]{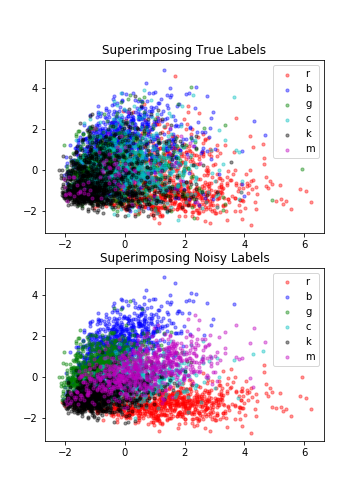}
		\caption*{Iteration 18}
	\end{subfigure}
	\caption{No noise model stacked}
	\label{fig::tsne_Wonm}
\end{subfigure}
	\caption{tSNE visualization of the last layer activations of a base network before softmax for Trec Dataset with $50\%$ corrupted labels; First row in (a) when the corresponding true labels are superimposed on the tSNE data points; Second row in (a) when the noisy labels are superimposed onto the tSNE data points.}
	\label{fig:tsne_results}
\end{figure*}

In order to further understand the noise model, we first train the base model and the proposed model on noisy labels. Afterward, we collect the last fully-connected layer's activations for all the training samples and treat them as the learned feature representation of the input sentence. We get two different sets of feature representations, one corresponding to the base model (\emph{TRB}), and the other corresponding to the proposed model (\emph{TRPr}). Given these learned feature representations -- the artificially injected noisy labels and the true labels of the training data -- we learn two different SVMs for each model, with and without noise. For the base model, for both SVMs, we use TRB representation as inputs and train the first SVM with the true labels as targets and the second SVM with the unreliable labels as targets. Similarly, we train two SVMs for the proposed model. After training, we evaluate the performance of all the learned SVMs on clean test data in Table \ref{table::svm_performance}, where the 1st column represents the corresponding model performance, ``Noisy" and ``True" column represents the SVM performance when trained on noisy and clean labels, respectively. We run these experiments for different datasets with different label noise.

The SVM, trained on TRB and noisy labels, is very close to the base model performance (\ref{table::svm_performance}). This suggests that the base model is just fitting the noisy labels. On the other hand, when we train an SVM on the TRPr representations with true labels as targets, the SVM achieves the proposed model performance. This means that the proposed approach helps the base model to learn better feature representations even with the noisy targets, which suggest that this noise model is learning a label denoising operator.

We analyze the representation of training samples in feature domain by plotting the tSNE embeddings \cite{van2014accelerating} of the TRB and TRPr. For brevity, we plot the t-SNE visualizations for trec dataset with $50\%$ label noise in Fig. \ref{fig:tsne_results} .

For each network, we show two different t-SNE plots. For example in Fig. \ref{fig::tsne_Nmwregu} we plot two rows of tSNE embeddings for the proposed model. In the first row of Fig. \ref{fig::tsne_Nmwregu}, each training sample is represented by its corresponding true label, while in the second row (the noisy label plot) each training sample is represented by its corresponding noisy label. We observe that, as the learning process progresses, the noise model helps the base model to cluster the training samples in the feature domain. With each iteration, we can see the formation of clusters in Row 1. However, in Row 2, when the noisy labels are superimposed, the clusters are not well separated. This means that the noise model denoises the labels and presents the true labels to the base network to learn. 

In Fig. \ref{fig::tsne_Wonm}, we plot two rows of tSNE embeddings of the TRB representations. It seems that the network directly learns the noisy labels. This provides further evidence to support \cite{zhang2016understanding}'s finding that the deep network memorizes data without knowing of true labels. In Row 2 of Fig. \ref{fig::tsne_Wonm}, we can observe that the network learns noisy features representations which can be well clustered according to given noisy labels.


\section{Conclusion}
\label{sec:conclusion}
In this work, we propose a framework to enable a DNN to learn better sentence representations in the presence of label noise for text classification tasks. To model the label noise, we append a non-linear noise model on top of the base CNN architecture. With proper initialization and regularization, the noise model is able to absorb most of the label noise and helps the base model to learn better sentence representations. 

\bibliography{ref_noise}
\bibliographystyle{acl_natbib}

\end{document}